\documentclass[11pt]{article}

\usepackage[]{acl}

\usepackage{times}
\usepackage{latexsym}

\usepackage[T1]{fontenc}

\usepackage[utf8]{inputenc}

\usepackage{microtype}

\usepackage{color}
\usepackage[most]{tcolorbox}
\usepackage{xcolor}
\usepackage{fontawesome5}
\usepackage{tabularx}
\usepackage{tabu}
\usepackage{booktabs}
\usepackage{multirow}
\usepackage{placeins}

\usepackage{graphicx} 
\usepackage{float}
\usepackage{subfigure} 
\usepackage{amssymb}
\providecommand{\mathbbm}[1]{\mathbb{#1}}
\usepackage[table]{xcolor}

\usepackage{hyperref}
\usepackage{tablefootnote}
\usepackage{xspace}

\usepackage{float}
\usepackage{amsmath}
\usepackage{bm}
\usepackage{algorithmicx}
\usepackage{algorithm}
\usepackage{algpseudocode}

\usepackage{arydshln}

\usepackage{amssymb}
\usepackage{pifont}
\usepackage{enumitem}

\newcommand{\traindata}{\textsc{TexOCR}-Train\xspace}
\newcommand{\model}{\textsc{TexOCR}\xspace}
\newcommand{\evaldata}{\textsc{TexOCR}-Bench\xspace}
\newcommand{\eg}{\hbox{\emph{e.g.,}}\xspace}

\newcommand{\nexample}{2,135\xspace}
\newcommand{\nmodel}{21\xspace}
\newcommand{\hflogo}{\raisebox{-0.15em}{\includegraphics[height=1em]{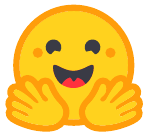}}}
\newcommand{\ghlogo}{\faGithub}

\usepackage{graphicx}

\title{\model: Advancing Document OCR Models for Compilable Page-to-LaTeX Reconstruction}

\definecolor{YaleBlue}{RGB}{0, 53, 107}
\definecolor{UNCBlue}{RGB}{75, 156, 211}
\definecolor{ZJBlue}{RGB}{1, 126, 199}

\newcommand{\Yale}{\hspace{.1em}^{\textcolor{YaleBlue}{\boldsymbol{Y}}}}
\newcommand{\ZJ}{\hspace{.1em}^{\textcolor{ZJBlue}{\boldsymbol{Z}}}}

\author{
Chengye Wang$\ZJ$\quad
Lin Fu$\ZJ$ \quad
Zexi Kuang$\Yale$ \quad
Yilun Zhao$\Yale$ \\[7pt]
$\Yale$Yale University \qquad $\ZJ$Zhejiang University \\[7pt]
\begin{tabular}{clcl}
\hflogo & \href{https://huggingface.co/collections/chengyewang/texocr}{Data \& Models} &
\ghlogo & \href{https://github.com/QDRhhhh/TexOCR}{Code}
\end{tabular}
}

\begin{document}
\maketitle
\begin{abstract}
Existing document OCR largely targets plain text or Markdown, discarding the structural and executable properties that make LaTeX essential for scientific publishing. We study page-level reconstruction of scientific PDFs into compilable LaTeX and introduce \evaldata, a benchmark, and \traindata, a large-scale training corpus, for this task. \evaldata features a multi-dimensional evaluation suite that jointly assesses transcription fidelity, structural faithfulness, and end-to-end compilability. Leveraging \traindata, we train a 2B-parameter model, \model, using supervised fine-tuning (SFT) and reinforcement learning (RL) with verifiable rewards derived from LaTeX unit tests that directly enforce compilability and referential integrity. Experiments across \nmodel frontier models on \evaldata show that existing systems frequently violate key document invariants, including consistent section structure, correct float placement, and valid label–reference links, which undermines compilation reliability and downstream usability. Our analysis further reveals that RL with verifiable rewards yields consistent improvements over SFT alone, particularly on structural and compilation metrics.

\end{abstract}
\section{Introduction}
Scientific knowledge remains predominantly distributed as PDF, yet most downstream workflows depend on editable, structure-preserving representations that support retrieval, reuse, and reproducible publishing (for example, compilation-ready LaTeX)~\cite{lo-etal-2020-s2orc}. Traditional OCR systems have historically been modular, combining document analysis, layout parsing, text recognition, and post-processing rules~\cite{pdfa2015,wick2018calamarihighperformancetensorflowbased,saha2010houghtransformbasedtechnique}. More recently, OCR has shifted toward multimodal large language models (MLLMs) that can transcribe directly from document images~\cite{greif2025multimodal,inoue2025contextindep,chen2025ocean,wang2025infinityparserlayoutaware,DianJin-OCR-R1}, raising the ceiling on end-to-end recognition quality and enabling broader general-purpose and OCR-specialized systems.

\begin{figure}[!t]
    \centering
    \includegraphics[width=\columnwidth]{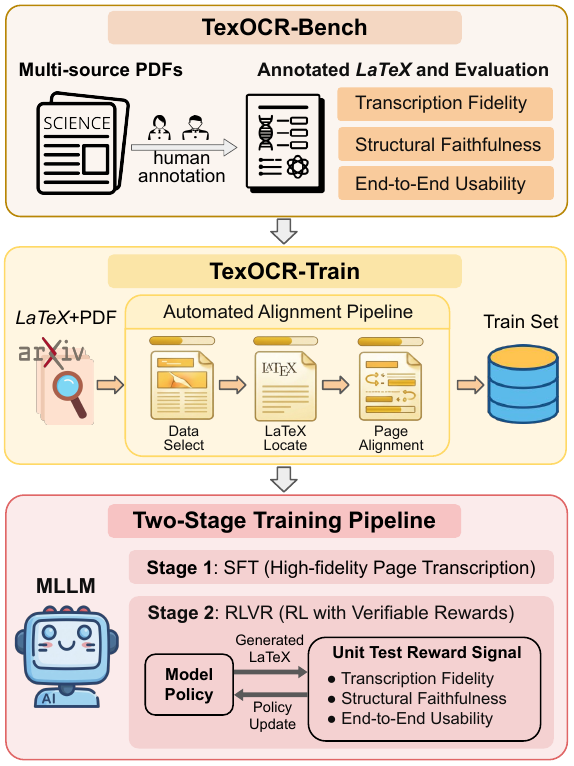}
    \caption{Overview of our work, covering data synthesis, model training, and multi-dimensional evaluation.}
    \label{fig:overview}
\end{figure}

Despite these advances, reliable page-level reconstruction of scientific papers into executable LaTeX remains under-served. 
Existing work primarily linearizes PDFs into plain text or Markdown~\cite{li-etal-2025-readoc,heakl-etal-2025-kitab}. When LaTeX is considered, it is typically framed as a localized conversion task for specific regions, most commonly math equations and tables~\cite{olmOCR,jiang2025latte,ling2025table2latex}.
These approaches under-specify the global constraints that determine whether a reconstructed project is actually usable: cross-page structure, stable float placement, consistent numbering, and reference integrity across citations, equations, figures, tables, and sections. As a result, even when surface transcription quality appears high, reconstructed LaTeX sources can remain brittle. Minor errors that are inconsequential in plain text, such as a missing brace, an unescaped character, an incorrect environment boundary, or a mismatched label, can render the document uncompilable or silently corrupt semantics by redirecting references and numbering.

To bridge this gap, we introduce \evaldata, a new evaluation benchmark for end-to-end, page-level LaTeX reconstruction with an explicit emphasis on transcription accuracy, executability, and structural fidelity. \evaldata contains \nexample expert-annotated examples spanning diverse document types and technical domains.
To move beyond surface transcription scores, we design a comprehensive evaluation protocol that jointly measures transcription fidelity, structural faithfulness, and end-to-end usability. The protocol includes component-level checks (\eg text, table, and formula accuracy), structure-aware checks (\eg section accuracy, citation coverage, and figure/table reference validity), and a compilation success rate computed by compiling the reconstructed project without manual intervention.

To facilitate future research, we release a large-scale training dataset, \traindata, and a training recipe tailored to compilable LaTeX reconstruction rather than generic OCR. We collect and process arXiv LaTeX source archives paired with PDFs at scale and construct page-aligned supervision, resulting in 57K papers and 404K (page image, LaTeX) training pairs. Building on this corpus, we train \model on Qwen3-VL-2B~\cite{qwen3technicalreport} with SFT for high-fidelity page transcription, and then further optimize it using reinforcement learning with verifiable rewards (RLVR). To adopt reliable reward functions for effective LaTeX-focused model development, we design binary unit tests that mirror the nine evaluation metrics in \evaldata, spanning transcription fidelity, structural faithfulness, and end-to-end usability.

We conduct a systematic evaluation of \nmodel frontier MLLMs and OCR models on \evaldata. Our results show that \evaldata poses a substantial challenge: even the best-performing model, GPT-5.3, achieves only 78.5 accuracy. While OCR systems such as olmOCR2~\cite{olmOCR2} and DeepSeek-OCR~\cite{wei2025deepseek} perform competitively on previous PDF-to-Markdown benchmarks, they degrade markedly on \evaldata, indicating that reconstructing compilable, structure-consistent LaTeX requires capabilities beyond high-quality transcription. 
Among open-source models, \model attains state-of-the-art performance on \evaldata. Moreover, we find that our RLVR scheme yields consistent improvements over SFT across both component-level and structure-aware metrics, driven by verifiable unit-test rewards that directly target compilation and referential integrity. Qualitative analysis further suggests that RLVR encourages the model to internalize LaTeX ``invariants'' (\eg balanced delimiters, well-formed environments, and stable label–reference alignment), reducing brittle failure modes that otherwise prevent successful compilation or silently corrupt document semantics.

Our main contributions are summarized below:
\begin{itemize}[leftmargin=*]
\itemsep0em
\item We introduce \evaldata, a new benchmark for end-to-end, page-level reconstruction of scientific PDFs into compilable LaTeX, with \nexample expert-annotated examples spanning diverse document types and technical domains.
\item We propose a comprehensive evaluation protocol that goes beyond surface similarity by jointly measuring component-level transcription fidelity, structure-aware consistency, and end-to-end usability via zero-touch compilation success.
\item We release \traindata, a large-scale training corpus, along with a training recipe that combines SFT and RLVR methods. Using this recipe, we train a 2B model, \model.
\item We conduct a systematic evaluation of \nmodel frontier models on \evaldata, showing that \model achieves the strongest performance among open-source baselines and that our RLVR design delivers consistent improvements on both component- and structure-level metrics.
\end{itemize}
\section{Related Works}
\paragraph{OCR Evaluation Benchmarks.}
Recent benchmarks evaluate document OCR along two complementary axes. (1) \emph{End-to-end conversion from PDF to Markdown or plain text}: READoc~\cite{li-etal-2025-readoc} and OmniDocBench~\cite{ouyang2025omnidocbenchbenchmarkingdiversepdf} define unified protocols for realistic multi-page extraction across diverse PDFs, while olmOCR-Bench~\cite{olmOCR2} adds pass or fail unit tests that check key document properties such as text presence, reading order, tables, formulas, and baseline functionality. (2) \emph{Structured element transcription to LaTeX for tables and formulas}: Table2LaTeX-RL~\cite{ling2025table2latex} evaluates table reconstruction with layout-consistent scoring and render-based checks, and CMER-Bench~\cite{bai2025complexmathematicalexpressionrecognition} benchmarks formula transcription with difficulty stratification. Across these settings, evaluation typically combines string or token similarity, consistency checks, and unit-test pass rates. In contrast, we introduce a new OCR task and a more comprehensive protocol that complements component metrics with structural checks and a zero-touch compilation test, yielding a stricter measure of end-to-end usability.

\paragraph{Document OCR Methods.}
Document OCR has increasingly shifted from modular pipelines to MLLMs that transcribe directly from document images. Early studies showed that encoder--decoder MLLMs can convert scientific PDFs into plain text without explicit layout supervision~\cite{blecher2023,wei2024generalocrtheoryocr}. Recent approaches include end-to-end generation of linearized text or markup~\cite{nassar2025smoldoclingultracompactvisionlanguagemodel,liu-etal-2025-points,li2025dotsocrmultilingualdocumentlayout,Duan2026GLMOCRTR,Dong2026QianfanOCRAU,Wu2026FireRedOCRTR,Zheng2026MultimodalOP,Taghadouini2026LightOnOCRA1} as well as hybrid systems that use MLLMs as recognition backbones within conventional document processing frameworks~\cite{wei2025deepseek,PaddleOCR-VL,Wang2026MinerU25ProPT,Wei2026DeepSeekOCR2V}. Earlier systems mainly relied on large-scale SFT on paired document images and text, such as Nanonets-OCR2~\cite{Nanonets-OCR2} and olmOCR~\cite{olmOCR}, whereas newer work increasingly adopts RL with task-specific or verifiable rewards to improve consistency on tables, formulas, and reading order like DianJin-OCR-R1~\cite{DianJin-OCR-R1}, olmOCR2~\cite{olmOCR2}. In contrast, we target compilable page-to-LaTeX reconstruction and introduce benchmarking and training signals that directly optimize executability and global LaTeX invariants beyond surface transcription.

\section{\evaldata Benchmark}
This section first discusses the evaluation suite of \evaldata, and then details the benchmark construction process.
\subsection{Evaluation Suite}
\label{app:eval_metric}
To comprehensively assess both OCR quality and LaTeX reconstruction fidelity, we organize nine metrics along three complementary dimensions:

\paragraph{Transcription Fidelity.} It measures fine-grained content recovery from the page image.
\textbf{Complex Text Preservation (CTP)} extracts one plain-text sentence per section (excluding LaTeX-specific commands) from the reference source and checks whether each sentence appears in the generated output, with strict matching on case sensitivity, punctuation, and character-level accuracy.
\textbf{Formula Accuracy (FA)} removes labels, environments, and layout-related noise from mathematical expressions, applies expression normalization, and compares formulas via string-based sequence matching.
\textbf{Table Accuracy (TA)} extracts and normalizes numerical entries in tables, then matches generated tables against the ground truth based on (i) the overlap ratio of numerical entries and (ii) the hit rate of \emph{unique numbers}.

\paragraph{Structural Faithfulness.} It checks document-level structure and cross-reference consistency. Specifically, 
\textbf{Section Accuracy (SA)} verifies whether all section titles are correctly recovered and aligned with the original document structure, including both title text matching and hierarchical correctness (\eg proper restoration of \verb|\section|).
\textbf{Citation Coverage (CC)} extracts bibliography information from the aggregated LaTeX and evaluates whether each in-text citation is correctly generated, checking the correctness of \verb|\cite{}| keys, their positions in the text, and the presence of missing or misrecognized citations.
\textbf{Reference Validity (RV)} checks whether figure and table references conform to valid LaTeX syntax and semantics, detecting undefined references and verifying that commands like \verb|\ref{fig:figure_x}| correctly resolve to existing labels.

\begin{table*}[t]
\centering
\small
\setlength{\tabcolsep}{6pt}
\begin{tabular}{lcccccc}
\toprule
\textbf{Set} & \textbf{\#Documents} & \textbf{\#Figures (Conv.)} & \textbf{\#Images} & \textbf{\#Tables} & \textbf{\#Formulas} & \textbf{Average LaTeX Length} \\
\midrule
\traindata & 57K & 404K & 181K & 231K & 488K & 31.8K \\
\evaldata  & 2K & 14.5K & 7.5K & 13.9K & 31K & 39.1K \\
\bottomrule
\end{tabular}
\caption{Dataset statistics for \traindata and \evaldata. \#Figures (Conv.) denotes figures converted from PDF pages. LaTeX Length reports the average length of the corresponding LaTeX source.}
\label{tab:data_statistics}
\end{table*}

\paragraph{End-to-End Usability.} It evaluates overall output quality and executability.
\textbf{Document-Level Similarity (DS)} concatenates the original LaTeX source into a single reference file and computes the normalized character-level edit distance between the generated LaTeX and the ground truth as $\mathrm{Similarity} = 1 - \mathrm{Levenshtein}(A, B)/\max(|A|, |B|)$, where $A$ and $B$ denote the reference and generated texts.
\textbf{Baseline} identifies severe page-level generation failures through basic usability checks, such as large-scale truncation, token loss, or abnormal characters.
\textbf{Compilation Success Rate (CSR)} integrates all generated LaTeX fragments into a complete project and compiles it using a standard LaTeX pipeline without manual intervention, reflecting overall usability, engineering completeness, and practical readiness of the output.

\subsection{Sourcing Documents and Creating Tests}
\label{sec:Sourcing_Documents}
Given the evaluation suite above, we construct \evaldata in two stages: sourcing a diverse pool of PDF documents, and expert annotation followed by test-set assembly.

\paragraph{Data Sources.}
We collect source documents in PDF format from multiple heterogeneous sources, covering both contemporary scientific articles and scanned materials~\cite{olmOCR}. Specifically, the data sources include:
(1) \textbf{arXiv} from which we collect scientific papers with publicly available PDF sources;
(2) \textbf{Public-domain mathematics textbooks}, where we randomly sample pages from each document to capture diverse mathematical notation and formatting styles;
(3) \textbf{The Library of Congress digital archives}, from which we sample historical letters and typewritten documents that come with existing human transcriptions;
(4) \textbf{The Internet Archive}, which serves as a large repository of public-domain scanned documents, including journals, government reports, and technical manuals.

\paragraph{Document Selection and Annotation.}
From this pool, we select 2{,}135 documents of moderate length with rich LaTeX-relevant structures, ensuring comprehensive coverage of the evaluation dimensions introduced above, such as figures, tables, mathematical formulas, and hierarchical section organizations. These selected documents constitute the initial dataset. Given the selected PDFs, we assign annotation tasks to human annotators, who are instructed to transcribe the corresponding content into LaTeX format following the guidelines detailed in Appendix~\ref{app:ocr_benchmark_construction}. We then construct the test set by consolidating each annotated LaTeX project into a single \texttt{.tex} file using rule-based heuristics. From the merged source, we extract figures, tables, equations, and section headings with lightweight parsers, which serve as ground-truth annotations for metric computation. Detailed implementations of all evaluation metrics are provided in Appendix~\ref{app:metric_details}.

\section{\traindata Construction}
To enable training at scale, we release \traindata. The following subsections detail the automated data construction process.
\subsection{Paper Collection}
To construct our benchmark, we programmatically collect official \texttt{.tar.gz} LaTeX sources and PDFs from arXiv, covering papers submitted between January~2022 and October~2025. For each paper, we resolve cross-file dependencies and merge all reachable \texttt{.tex} files into a single canonical source while preserving document order and file provenance. On this unified representation, we apply a deterministic rule-based parser to recover document structure, including section hierarchy as well as figure and table constructs. These normalized structural annotations provide standardized inputs and supervision for downstream benchmark tasks.

\subsection{Train Set Design}
To construct a high-quality training dataset for PDF-to-LaTeX conversion, we design a structured alignment and labeling pipeline based on the collected corpus. 
Each document is segmented into a set of single-page screenshots and training samples are organized as paired instances of (page image, textual supervision).

\paragraph{Page--Body Alignment.}
We use reference markers to find where the references start. Everything before that is treated as the main body, and only the first reference page is kept as the reference section. For pages in the main body, we use GPT5-mini~\cite{openai_gpt5_2025} to detect begin/end token boundaries of the page content and align them to the crawled LaTeX source.

\paragraph{Float Placement.}
A key challenge is that the rendered PDF layout does not always align with the linear order of the LaTeX source: figures and tables defined within a given section may appear on different pages due to float placement or pagination. To resolve this mismatch, we use pdf2figure~\cite{pdf2figure} to detect and localize figures and tables across the entire PDF, and leverage the resulting global layout information to assign each element to its most appropriate page, yielding a consistent mapping between each page image and its complete LaTeX representation.

\paragraph{Bibliography Supervision.}
For the reference section, our supervision target is not the LaTeX-rendered bibliography list but rather BibTeX entries. Accordingly, we use GPT-based OCR outputs as labels for reference pages. For the first page containing references, we apply a hybrid labeling strategy: non-reference regions are supervised with standard LaTeX, whereas reference regions are converted into the corresponding BibTeX format. This design encourages the model to learn a structured switch from body LaTeX generation to bibliography entry generation near the end of the document. Then we apply this pipeline to all papers, resulting in 57K papers and 404K training pairs (page image, LaTeX/BibTeX text). 

\section{Two-Stage Training Pipeline}
\label{sec:training}

We train our PDF-to-LaTeX model in two stages: SFT for faithful page-level transcription, followed by RLVR to target failure modes such as formula rendering and table structure that are poorly captured by token-level likelihood.

\subsection{Stage I: SFT}
We initialize from a strong instruction-tuned multimodal backbone and fine-tune it on our page-aligned pairs
$(x, y)$, where $x$ is a single-page PDF screenshot and $y$ is the corresponding LaTeX (or BibTeX for reference regions).
Given an instruction prompt $p$ (describing the desired output format and conventions), the model is trained with standard
next-token prediction:
\begin{equation}
\small
\mathcal{L}_{\text{SFT}}
~=~
-\mathbb{E}_{(x,y)\sim\mathcal{D}}
\left[
\sum_{t=1}^{|y|}
\log \pi_{\theta}(y_t \mid x, p, y_{<t})
\right].
\end{equation}
For the first reference page, we use mixed supervision: body regions are supervised in LaTeX, while detected reference regions
are supervised in BibTeX entries, encouraging a clean transition from body generation to bibliography generation near the end of a paper.

\begin{table*}[!t]
\centering
\footnotesize
\renewcommand{\arraystretch}{1.1}
\resizebox{\linewidth}{!}{
\begin{tabular}{l|cccc|cccc|cccc|c}
\toprule
\multirow{2}{*}{\textbf{Method}}
& \multicolumn{4}{c|}{\textbf{Structural Faithfulness}}
& \multicolumn{4}{c|}{\textbf{End-to-End Usability}}
& \multicolumn{4}{c|}{\textbf{Transcription Fidelity}}
& \multirow{2}{*}{\textbf{Overall}} \\
& SA & CC & RV & Avg
& DS & Baseline & CSR & Avg
& CTP & FA & TA & Avg
&  \\
\midrule
GPT-5.3 & \textbf{83.4} & \textbf{86.0} & \underline{65.2} & \underline{78.2} & \textbf{72.4} & \textbf{98.8} & \textbf{82.7} & \textbf{84.6} & 69.2 & 57.9 & \textbf{91.1} & 72.7 & 78.5 \\
\rowcolor{gray!20}
\textbf{\model} (SFT + RLVR) & 76.7 & \underline{85.9} & \textbf{86.8} & \textbf{83.1} & 64.7 & 95.2 & 45.2 & 68.4 & 72.8 & 58.4 & \underline{89.2} & 73.5 & 75.0 \\
\rowcolor{gray!20}
\textbf{\model} (SFT) & 73.5 & 74.5 & 74.1 & 74.0 & 61.8 & 91.8 & 44.3 & 66.0 & 69.0 & 53.3 & 88.0 & 70.1 & 70.0 \\
Qwen3-VL-32B & 61.0 & 62.6 & 42.9 & 55.5 & 67.0 & 98.3 & 58.9 & \underline{74.7} & 75.7 & \underline{63.8} & 88.9 & \textbf{76.1} & 68.8 \\
Qwen3-VL-8B & 72.9 & 35.6 & 9.7 & 39.4 & 66.1 & \underline{98.7} & 59.0 & 74.6 & \textbf{77.2} & 62.2 & 77.0 & 72.1 & 62.2 \\
Mistral-Small-3.1-24B & 68.9 & 43.2 & 44.6 & 52.2 & 60.4 & 98.4 & 39.1 & 66.0 & 62.1 & 47.4 & 79.3 & 62.9 & 60.4 \\
Infinity-Parser-7B & \underline{77.5} & 3.7 & 17.6 & 32.9 & \underline{68.9} & 96.5 & 44.5 & 70.0 & \underline{76.8} & \textbf{66.1} & 80.8 & \underline{74.6} & 59.1 \\
Qwen2.5-VL-32B & 68.0 & 15.9 & 8.6 & 30.8 & 62.1 & 96.2 & 55.8 & 71.4 & 67.6 & 62.7 & 81.9 & 70.7 & 57.6 \\
Qwen3-VL-4B & 64.2 & 17.3 & 8.5 & 30.0 & 63.1 & 98.2 & 58.4 & 73.2 & 69.1 & 58.7 & 52.6 & 60.1 & 54.5 \\
Qwen3-VL-2B & 49.6 & 21.8 & 1.5 & 24.3 & 50.3 & 97.9 & 57.4 & 68.5 & 67.3 & 55.2 & 68.8 & 63.8 & 52.2 \\
Qwen2.5-VL-7B & 53.9 & 4.8 & 9.2 & 22.6 & 52.3 & 94.6 & 58.2 & 68.4 & 56.8 & 53.2 & 51.1 & 53.7 & 48.2 \\
olmOCR-2-7B & 43.7 & 0.5 & 0.2 & 14.8 & 65.9 & 96.1 & 36.5 & 66.2 & 74.4 & 60.5 & 49.3 & 61.4 & 47.5 \\
Pixtral-12B & 52.1 & 4.8 & 2.1 & 19.7 & 50.8 & 92.7 & \underline{64.5} & 69.3 & 47.9 & 42.7 & 62.8 & 51.1 & 46.7 \\
InternVL3-38B & 58.3 & 21.1 & 10.4 & 29.9 & 40.8 & 92.3 & 39.6 & 57.6 & 26.3 & 38.2 & 61.3 & 41.9 & 43.1 \\
InternVL3-8B & 51.9 & 17.0 & 3.6 & 24.2 & 49.1 & 93.1 & 35.1 & 59.1 & 46.6 & 37.6 & 50.4 & 44.9 & 42.7 \\
InternVL2.5-38B & 54.6 & 13.0 & 8.1 & 25.2 & 45.1 & 88.9 & 42.7 & 58.9 & 36.9 & 35.0 & 43.1 & 38.3 & 40.8 \\
Qwen2.5-VL-3B & 52.2 & 1.3 & 0.9 & 18.1 & 43.2 & 76.9 & 39.5 & 53.2 & 43.2 & 43.8 & 45.2 & 44.1 & 38.5 \\
DeepSeek-OCR & 4.1 & 0.3 & 0.0 & 1.5 & 33.8 & 94.7 & 50.1 & 59.5 & 44.3 & 48.3 & 2.0 & 31.5 & 30.8 \\
Phi-4-multimodal & 18.6 & 7.7 & 6.2 & 10.8 & 19.2 & 98.3 & 56.8 & 58.1 & 1.5 & 26.6 & 5.4 & 11.2 & 26.7 \\
InternVL2.5-8B & 35.0 & 5.8 & 2.0 & 14.3 & 26.2 & 73.3 & 36.8 & 45.4 & 7.4 & 28.1 & 20.8 & 18.8 & 26.2 \\
LLaVA-OneVision & 26.4 & 2.2 & 0.6 & 9.7 & 22.7 & 65.4 & 46.1 & 44.7 & 6.1 & 26.8 & 12.6 & 15.2 & 23.2 \\
InternVL2-8B & 36.9 & 5.1 & 1.6 & 14.5 & 24.4 & 61.5 & 24.4 & 36.8 & 6.0 & 28.0 & 20.8 & 18.3 & 23.2 \\
Phi-3.5-vision & 22.5 & 1.2 & 1.1 & 8.3 & 15.6 & 87.7 & 44.9 & 49.4 & 2.5 & 26.6 & 4.8 & 11.3 & 22.9 \\
\bottomrule
\end{tabular}
}
\caption{Main results on \evaldata, reporting scores across three metric groups along with the Overall score: \textbf{Structural Faithfulness} --- Section Accuracy (SA), Citation Coverage (CC), and Reference Validity (RV); \textbf{End-to-End Usability} --- Document-Level Similarity (DS), page-level baseline sanity check (Baseline), and Compilation Success Rate (CSR); and \textbf{Transcription Fidelity} --- Complex Text Preservation (CTP), Formula Accuracy (FA), and Table Accuracy (TA). The best scores per column are in \textbf{bold} and second-best are \underline{underlined}.}
\label{tab:main_result}
\end{table*}

\subsection{Stage II: RLVR with Unit-Test Reward}
While SFT teaches the model to imitate page labels, it does not directly optimize for \emph{functional} correctness (e.g., whether a formula
renders correctly, whether a table preserves numeric content and structure, or whether cross-references resolve). Inspired by olmOCR2~\cite{olmOCR2}, which leverages unit-test pass rates as a verifiable training signal for document OCR, we therefore apply RLVR on
single-page inputs. For each page $x$, we sample a group of $K$ completions $\{y^{(k)}\}_{k=1}^{K}\sim\pi_{\theta}(\cdot\mid x,p)$ and score each
completion using a suite of automatically constructed \emph{binary unit tests}. The page-level reward is the fraction of unit tests that pass:
\begin{equation}
\small
R(x, y)
~=~
\frac{1}{|\mathcal{T}(x)|}
\sum_{\tau \in \mathcal{T}(x)}
\mathbbm{1}\!\left[\tau(y)=\texttt{pass}\right]
\in [0,1],
\end{equation}
where $\mathcal{T}(x)$ denotes the test set instantiated for page $x$ from its aligned ground truth and auxiliary layout signals.
We optimize the policy with a group-based RL objective (e.g., GRPO-style updates) using within-group normalization for variance reduction, and a
KL penalty to keep the learned policy close to the SFT reference policy $\pi_{\text{ref}}$:
\begin{equation}
\small
\begin{aligned}
\max_{\theta}\;
\mathbb{E}_{x}\left[
\frac{1}{K}\sum_{k=1}^{K}
\Big(
\hat{A}^{(k)} \cdot \log \pi_{\theta}(y^{(k)} \mid x,p)
\Big)
\right]
\\
-\;
\beta\,\mathbb{E}_{x}\!\left[\mathrm{KL}\!\left(\pi_{\theta}(\cdot\!\mid x,p)\,\|\,\pi_{\text{ref}}(\cdot\!\mid x,p)\right)\right].
\end{aligned}
\end{equation}
where $\hat{A}^{(k)}$ is a group-relative advantage derived from the rewards $\{R(x,y^{(k)})\}_{k=1}^{K}$.

\paragraph{Unit-Test Reward Design.}
We construct the page-level verifiable reward by adapting the full \evaldata evaluation suite (see Section~\ref{app:eval_metric}) to single-page scoring. For each page, we reformulate all \textbf{nine metrics} into deterministic binary (pass/fail) unit tests covering the three metric groups:
(i) \emph{transcription-fidelity tests} check faithful content recovery, including complex text preservation on extracted plain-text anchor sentences, table accuracy via numerical-entry overlap, and formula accuracy via token-level equivalence after normalization;
(ii) \emph{structural-faithfulness tests} probe document-level invariants, including section-hierarchy consistency, citation-key validity against the page's bibliography, and label--reference resolution for figures and tables;
(iii) \emph{end-to-end usability tests} target executable quality through a page-level sanity check that rejects empty, ill-formed, or degenerately repetitive outputs, a document-similarity test based on normalized edit distance, and a compilation probe that wraps the generated snippet in a minimal preamble and invokes a standard LaTeX compiler.
For continuous metrics, we binarize by thresholding so that each test produces a clean pass/fail outcome; the scalar reward is then the fraction of tests passed.

\begin{table*}[t]
\centering
\small
\setlength{\tabcolsep}{6pt}
\begin{tabular}{p{0.25\linewidth} p{0.70\linewidth}}
\toprule
\textbf{Error Type} & \textbf{Description} \\
\midrule
Paragraph Truncation and Missing Text &
Partial paragraphs are often omitted at page beginnings or endings, especially near page boundaries or section transitions, leading to systematic loss of textual fragments. \\
\midrule
Mathematical Formula Mismatches &
Formulas frequently contain incorrect symbols, missing or mismatched delimiters, or improper inline/display classification, resulting in invalid or non-compilable expressions. \\
\midrule
Table Structure Corruption &
Tables suffer from structural errors such as missing column separators, misaligned rows, or incorrect multi-line cell handling, which easily break table environments. \\
\midrule
Citation and Reference Errors &
Citations are sometimes omitted or generated with incorrect formats, causing unresolved \verb|\cite| or \verb|\ref| commands and broken cross-referencing. \\
\midrule
Compilation Failures &
Malformed formulas, corrupted tables, and reference errors often lead to LaTeX compilation failures, significantly limiting practical usability. \\
\bottomrule
\end{tabular}
\caption{Common error types observed in PDF-to-LaTeX generation and their typical manifestations.}
\label{tab:error_analysis}
\end{table*}

\section{Experiments}
\label{sec:experiments}
We benchmark a broad set of frontier models on \evaldata to assess the state of PDF-to-LaTeX reconstruction, analyze common failure modes, and validate the effectiveness of our training recipe through systematic ablations.
\subsection{Experimental Setup}
\label{sec:exp_setup}

We evaluate a broad set of contemporary MLLMs on \evaldata under a unified page-level inference protocol.
\paragraph{Models.}
We evaluate a diverse collection of frontier models that support both visual and textual inputs. On the open-source side, we benchmark \textbf{thirteen open-source model series}: InternVL-2/2.5/3~\cite{chen2024far,chen2024expanding}, Qwen2.5-VL~\cite{Qwen2.5-VL} and Qwen3-VL~\cite{qwen3technicalreport}, DeepSeek-OCR~\cite{wei2025deepseek}, olmOCR-2-7B~\cite{olmOCR2}, Infinity-Parser-7B~\cite{wang2025infinityparserlayoutaware}, Pixtral~\cite{agrawal2024pixtral12b}, Mistral-Small-3.1~\cite{mistral2025small31}, LLaVA-OneVision~\cite{li2024llavaonevisioneasyvisualtask}, Phi-3.5-Vision, and Phi-4-Multimodal~\cite{abdin2024phi3,microsoft2025phi4}. On the proprietary side, we include \textbf{OpenAI GPT-5.3} as the representative closed-source baseline. For open-source models, we run inference with vLLM~\cite{kwon2023efficient} to ensure efficient, reproducible decoding; the proprietary model is accessed via the official API using recommended default settings. 
The model prompt is provided in Appendix~\ref{app:infer_prompt}.
Detailed model configurations are provided in Appendix~\ref{app:model_info}.

\paragraph{\model Training.}
We fine-tune a \textbf{Qwen3-VL-2B} base model on our dataset using SFT. The model is trained with full-parameter updates for 1 epoch at a learning rate of $1e-5$. Each training sample pairs a rendered PDF page image with its corresponding LaTeX source.

\paragraph{Inference Protocol.}
We adopt a unified page-level inference protocol: each PDF page is rendered as a single image and processed independently, and the resulting LaTeX outputs are concatenated in document order for evaluation. This page-level setting is a necessary compromise---multi-image and merged multi-page alternatives we investigated suffer from cross-page interference and resolution loss, respectively, as detailed in Section~\ref{sec:inference_granularity}.

\paragraph{Evaluation Metrics.}
We report the full nine-metric suite introduced in Section~\ref{app:eval_metric}, organized along the three dimensions defined there. Unless otherwise specified, page-level outputs are first merged to form document-level predictions, and metrics are then computed at the document level and averaged across all documents.

\subsection{Main Findings}
Table~\ref{tab:main_result} reports results across all models, yielding three observations. First, proprietary MLLMs are the most consistent: they preserve complex body text and remain strong on formulas and tables, while many open-source models handle plain text well but degrade on syntax-sensitive tasks with malformed equations, broken table structures, and inconsistent reference formatting. Second, our two-stage training recipe is effective: SFT alone already delivers large gains over the Qwen3-VL-2B base across all three dimensions, and adding RLVR further improves aspects poorly captured by token-level likelihood---particularly structural faithfulness and end-to-end usability---while preserving SFT's transcription quality. Third, compilation success is a stricter usability indicator than recognition metrics alone: many generated projects still fail to compile without manual correction, typically due to unclosed math environments, malformed tables, and inconsistent \verb|\ref{...}| labels, revealing that current models lack sufficient syntactic robustness and project-level consistency for reliable downstream use.

\subsection{Error Analysis and Case Study}
\label{sec:error_analysis}

To understand failure modes beyond aggregate metrics, we conduct a targeted error analysis on our evaluation set. Our analysis reveals several recurring error patterns that directly affect the completeness, correctness, and compilability of the generated LaTeX. We summarize these errors in Table~\ref{tab:error_analysis} and illustrate each category with representative case studies in Appendix~\ref{sec:errorana}. A particularly extreme failure mode is exemplified by DeepSeek-OCR: regardless of how we modify the prompt, the model consistently produces Markdown-style linearized output rather than LaTeX, leaving virtually no valid section, citation, reference, or \verb|tabular| commands for the evaluator to score. This accounts for its near-zero scores on structural faithfulness and table accuracy in Table~\ref{tab:main_result}, despite competitive performance on coarser transcription checks.

\begin{table}[!t]
\centering
\small
\setlength{\tabcolsep}{6pt}
\resizebox{\columnwidth}{!}{%
\begin{tabular}{lccc}
\toprule
\textbf{Model} & \textbf{Single-Image} & \textbf{Multi-Image} & \textbf{Merged} \\
\midrule
Qwen3VL-2B & 52.2 & 39.1 & 36.9 \\
GPT-5.3    & 78.5 & 56.9 & 42.6 \\
\bottomrule
\end{tabular}%
}
\caption{Overall score on \evaldata under single-image, multi-image, and merged inference strategies.}
\label{tab:inference_strategy}
\end{table}

\begin{table*}[t]
\centering
\footnotesize
\resizebox{\linewidth}{!}{
\begin{tabular}{l|cccc|cccc|cccc|c}
\toprule
\multirow{2}{*}{\textbf{Method}} 
& \multicolumn{4}{c|}{\textbf{Structural Faithfulness}} 
& \multicolumn{4}{c|}{\textbf{End-to-End Usability}} 
& \multicolumn{4}{c|}{\textbf{Transcription Fidelity}} 
& \multirow{2}{*}{\textbf{Overall}} \\
& SA & CC & RV & Avg
& DS & Baseline & CSR & Avg
& CTP & FA & TA & Avg
& \\
\midrule
SFT+RLVR
& 76.7 & 85.9 & 86.8 & 83.1
& 64.7 & 95.2 & 45.2 & 68.4
& 72.8 & 58.4 & 89.2 & 73.5
& 75.0 \\

\quad $w.o.$ Transcription Fidelity
& 75.0 & 76.5 & 80.6 & 77.4
& 60.3 & 93.3 & 48.9 & 67.5
& 65.6 & 53.4 & 87.7 & 68.9
& 71.3 \\

\quad $w.o.$ Structural Faithfulness
& 74.8 & 69.4 & 75.1 & 73.1
& 61.6 & 93.5 & 46.4 & 67.2
& 68.6 & 54.6 & 88.5 & 70.6
& 70.3 \\

\quad $w.o.$ End-to-End Usability
& 75.2 & 72.5 & 77.6 & 75.1
& 60.2 & 93.0 & 46.3 & 66.5
& 67.7 & 54.2 & 88.3 & 70.1
& 70.5 \\

\bottomrule
\end{tabular}
}
\caption{Ablation of verifiable unit tests in RLVR, where removing a specific test leads to a clear degradation in corresponding evaluation metric, highlighting the targeted supervision provided by each unit-test reward.}
\label{tab:ablation2}
\end{table*}
\subsection{Effect of Inference Granularity}
\label{sec:inference_granularity}
We compare single-image inference with multi-image and merged multi-page inference strategies for PDF-to-LaTeX conversion. As shown in Table~\ref{tab:inference_strategy}, single-image inference consistently achieves better performance. Multi-image inference introduces cross-page interference that degrades visual–text alignment, while merging multiple pages into a single image leads to resolution loss and visual clutter, both of which harm structural accuracy. In contrast, single-image inference preserves page-level locality and visual clarity, resulting in more accurate and stable LaTeX generation.

\subsection{Ablation Study}
We conduct a set of controlled ablation studies to isolate the contributions of different training components and design choices in our framework.

\paragraph{Ablation of Verifiable Unit Tests.}
Next, we ablate individual categories of verifiable unit tests in RLVR to assess their impact. As shown in Table ~\ref{tab:ablation2}, removing a given test consistently degrades its corresponding evaluation metric (e.g., lower formula accuracy without formula tests), with little compensation from other metrics. This highlights the targeted role of unit-test rewards: each test explicitly supervises a specific functional capability, and its absence directly weakens that behavior. Overall, these results show that gains in structural faithfulness and end-to-end usability require explicit, verifiable objectives rather than emerging implicitly.

\paragraph{Effect of Group Size $K$.}
Finally, we analyze the effect of group size $K$ in RLVR for $K \in \{4, 8, 12, 16, 20, 24\}$. Larger $K$ consistently yields more stable performance across metrics, while small $K$ leads to higher variance and less reliable gains. Increasing $K$ reduces optimization noise and produces smoother improvements, indicating more accurate relative advantage estimates and stronger variance reduction. Overall, sufficiently large group sizes are crucial for stable training and for fully realizing the benefits of RLVR.

\begin{figure}[t]
    \centering
    \includegraphics[width=\columnwidth]{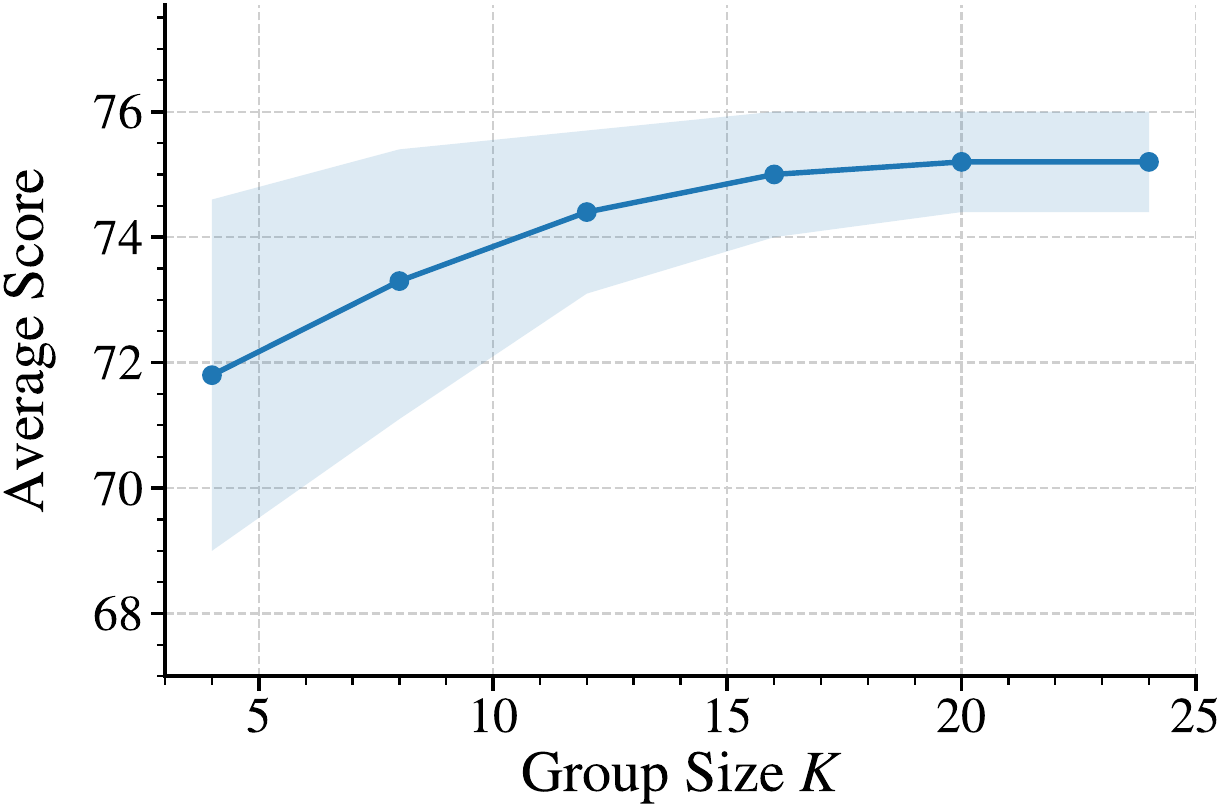}
    \caption{Effect of group size $K$ in RLVR. We report the average benchmark score as a function of $K$, with shaded regions indicating the standard deviation across evaluation samples.}
    \label{fig:group_size_k}
\end{figure}

\section{Conclusion}
This work targets a practical gap in PDF-to-LaTeX reconstruction: generating output that not only looks correct, but also compiles and preserves document structure reliably. We introduce \evaldata to evaluate executable, structure-faithful reconstruction at scale, and \traindata to enable training with page-aligned supervision. Using unit-test style, verifiable rewards, we show that optimizing directly for functional correctness reduces common failure modes such as broken environments, malformed tables, and invalid references. These resources and results establish a stronger baseline for developing OCR models that can recover scientific documents into usable LaTeX, and provide actionable insights for future advancements.

\section*{Limitation}
This work provides a first step toward reliable page-level reconstruction of scientific documents into compilable LaTeX. While our results demonstrate strong performance under a page-wise inference setting, the overall task is inherently document-centric. An important direction for future work is to move beyond isolated page-level reconstruction and develop more effective document-level approaches that better capture cross-page structure, global consistency, and long-range dependencies.

\bibliography{custom,anthology-1}

\appendix

\clearpage
\section{Implementation Details of Evaluation Metrics}
\label{app:metric_details}

This appendix provides a detailed description of the implementation of each evaluation metric used in our benchmark. All metrics are computed deterministically based on the generated LaTeX output and the corresponding ground-truth LaTeX source. Unless otherwise stated, evaluation is performed at the document level by aggregating page-level outputs in their original order.
\subsection{Complex Text Preservation}
\label{app:context}

To assess whether complex body text is preserved, we extract one representative plain-text sentence from each section of the ground-truth LaTeX source. Sentences containing LaTeX commands or environments are excluded to avoid trivial mismatches. For each extracted sentence, we perform an exact substring match against the concatenated generated LaTeX output. The context existence score is defined as the ratio of ground-truth sentences that are successfully recovered in the generated text.

\subsection{Document-Level Similarity}
\label{app:edit_distance}

We measure overall textual similarity between the generated LaTeX and the ground-truth source using normalized Levenshtein distance. Before comparison, BibTeX entries are removed from the generated output to avoid penalizing formatting differences in references. The similarity score is computed as
\[
\mathrm{Similarity} = 1 - \frac{\mathrm{Levenshtein}(A, B)}{\max(|A|, |B|)},
\]
where $A$ and $B$ denote the generated and reference LaTeX strings, respectively.

\subsection{Table Accuracy}
\label{app:table}

Table accuracy evaluates whether tabular content is correctly reconstructed. For each reference table, we identify the corresponding generated table on the same page. We extract numerical entries (including percentages) from the longest \verb|tabular| environment and normalize them into numeric tokens. Two tables are considered a match if they satisfy either:
(i) a high overlap ratio between numerical entries; or
(ii) a moderate overlap ratio combined with a high hit rate on unique numerical anchors (numbers that appear exactly once in the reference table).
This metric is binary per table. The final table accuracy is computed as the proportion of reference tables that are successfully matched.

\subsection{Formula Accuracy}
\label{app:equation}

Formula accuracy measures whether mathematical expressions are faithfully transcribed. We extract display math environments (e.g., \verb|equation| and \verb|eqnarray|) from both generated and reference LaTeX. Each generated formula is aligned to at most one reference formula using sequence similarity after normalization, which removes comments, layout commands (e.g., \verb|\left|, \verb|\right|), and whitespace. Alignment is accepted if the similarity exceeds a fixed threshold. For matched formula pairs, we further verify equivalence using token-level comparison: formulas are tokenized into LaTeX commands, variables, numbers, and operators, and are considered correct if the token sequences are identical or one is an ordered subsequence of the other. Formula accuracy is reported as the fraction of reference formulas that are correctly matched.

\subsection{Baseline Validity Check}
\label{app:baseline}

The baseline metric serves as a coarse-grained sanity check to detect severe page-level generation failures. For each generated page, we verify that the output satisfies a set of minimal usability constraints:
(i) the output is a non-empty string;
(ii) it contains at least one alphanumeric character;
(iii) it does not contain non-target language characters (e.g., CJK characters);
(iv) it does not contain emoji symbols; and
(v) it does not exhibit degenerate repetition patterns, defined as repeated trailing $n$-grams for any $n$ up to a fixed threshold.
A page is counted as valid if it passes all checks. The baseline score is computed as the fraction of valid pages among all pages in the document.

\subsection{Section Accuracy}
\label{app:section}

Section accuracy evaluates whether structural section headers are correctly recovered. We extract all occurrences of \verb|\section|, \verb|\subsection|, and \verb|\subsubsection| commands from both the generated and reference LaTeX. Section titles are normalized by removing numeric prefixes (e.g., ``3.2''). A predicted section is considered correct if its title matches a reference title under bidirectional substring inclusion. Each predicted section can be matched to at most one ground-truth section. We report precision as the fraction of predicted sections that are successfully matched.

\subsection{Citation Coverage}
\label{app:citation}

Citation coverage evaluates whether in-text citation commands are correctly generated. We extract all citation commands (e.g., \verb|\cite|, \verb|\citep|, \verb|\citet|) and their keys from both the generated output and the reference source. A generated citation key is considered valid if it either refers to a numeric citation index within the number of BibTeX entries or appears in the concatenated BibTeX content of the generated document. Citation precision is computed as the ratio of valid generated citations to the total number of ground-truth citations.

\subsection{Figure and Table Reference Validity}
\label{app:label}

To evaluate cross-reference correctness, we extract all \verb|\ref{...}| commands from both the generated and reference LaTeX. For each figure and table label defined in the reference, we count the number of times it is referenced. A label is considered correct if the generated output references it the same number of times as in the ground truth. Label precision is computed as the fraction of labels that satisfy this condition.

\clearpage
\onecolumn
\section{OCR Benchmark Construction}
\label{app:ocr_benchmark_construction}
\subsection{PDF2Latex Instruction}







\begin{figure}[h]
\small
\begin{tcolorbox}[
    colback=black!7.5!white,
    colframe=black!80!white,
    title=PDF-to-LaTeX Annotation Instructions (Data with LaTeX Sources)
]
You are given the original LaTeX source. Treat it as the source of truth. You are not rewriting content. Your task is to make it compile cleanly in our template while keeping the meaning unchanged.

Do not drop any visible text, equations, tables, captions, or partial sentences at page boundaries. Do not invent missing content. Keep math symbols, subscripts, superscripts, and indices exactly as they appear.

You may do light cleanup that does not change meaning, such as removing LaTeX comments, fixing whitespace, escaping special characters, and ensuring that braces and environments are balanced. Use standard environments such as \verb|\section|, \verb|\equation| or \verb|align|, \verb|table| and \verb|tabular|, and \verb|figure|. For figures, use an \verb|\includegraphics| placeholder and always include a caption and a label. For tables, keep the same column count and row breaks, and copy all numbers exactly.

Convert citations into \verb|\cite{}| and do not invent citation keys. Follow the dataset citation schema. If the source uses numeric citations, represent them as \verb|\cite{1,2}|. If the source uses author-year citations, map them to keys that match the provided BibTeX entries. Do not change which sentences the citations attach to.

Labels and references must be standardized and consistent. Use \verb|fig:figure_1|, \verb|fig:figure_2|, and so on for figures, and \verb|tab:table_1|, \verb|tab:table_2|, and so on for tables. Use \verb|eq:equation_1|, \verb|eq:equation_2|, and so on for displayed equations when an equation label is needed. Always reference them with \verb|\ref{}| or \verb|\eqref{}| using the same label you defined. If the text says ``Figure 3'' or ``Table 2'', make sure the label you assign matches that numbering, and update all references to match.
\end{tcolorbox}
\label{fig:latex_source_inst}
\end{figure}

\begin{figure}[h]
\small
\begin{tcolorbox}[
    colback=black!7.5!white,
    colframe=black!80!white,
    title=PDF-to-LaTeX Annotation Instructions (Data without LaTeX Sources)
]
You are given a PDF but not the original LaTeX source. Reconstruct the document as faithful and compilable LaTeX in our template. Within that constraint, match the original wording, math, and structure as closely as possible.

Avoid paraphrasing unless it is necessary to make the output compile. Use the same conventions as above for environments, figures, and tables. Figures should be represented with an \verb|\includegraphics| placeholder and must include a caption and a label. Tables should be rebuilt with the correct grid, separators, and row breaks, and you must copy all numeric entries exactly. If a table is extremely complex, prioritize compilability while preserving all numeric entries and any structure you can reliably recover.

For citations, keep them where they appear in the PDF. Convert them into \verb|\cite{}| without inventing keys and follow the dataset citation schema. Numeric citations should be written as \verb|\cite{1,2}|. Author-year citations should be mapped to keys that match the provided BibTeX entries. Do not move citations to different sentences.

For labels and cross-references, use a standardized scheme with explicit prefixes. Use \verb|fig:figure_1|, \verb|tab:table_1|, and \verb|eq:equation_1| style labels, and use \verb|\ref{}| or \verb|\eqref{}| consistently so cross-references resolve. Make sure the numbering is aligned with the PDF. If the PDF references ``Fig.~4'', you should label that figure as \verb|fig:figure_4| and ensure every mention points to \verb|\ref{fig:figure_4}|. If something in the PDF is unclear, do not guess. Keep it conservative, prioritize compilability, and preserve what is unambiguous.
\end{tcolorbox}
\label{fig:no_latex_source_inst}
\end{figure}

\newpage
\subsection{Annotator Information}
\begin{table}[h]
\centering
\small
\setlength{\tabcolsep}{4pt}
\begin{tabular}{lll}
\toprule
ID & Position & Disciplinary \\
\midrule
1  & Master Student     & Computer Science \\
2  & Master Student     & Engineering \\
3  & Master Student     & Healthcare \\
4  & Postdoc            & Nature Science \\
5  & Master Student     & Engineering \\
6  & Master Student     & Computer Science \\
7  & Master Student     & Healthcare \\
8  & PhD Student        & Nature Science \\
9  & Postdoc            & Engineering \\
10 & Master Student     & Nature Science \\
11 & PhD Student        & Healthcare \\
12 & Research Scientist & Computer Science \\
13 & Master Student     & Healthcare \\
14 & Master Student     & Engineering \\
15 & PhD Student        & Computer Science \\
\bottomrule
\end{tabular}
\caption{Information of annotators engaged in dataset construction. The average payment rate is 13 US dollars.}
\label{tab:candidate_profiles}
\end{table}

\clearpage
\onecolumn

\newpage
\section{Model Inference Prompt}
\label{app:infer_prompt}
\begin{figure}[h]
\centering
\footnotesize
\setlength{\fboxsep}{4pt}
\begin{tcolorbox}[
    colback=black!7.5!white,
    colframe=black!80!white,
    title=Model Inference Prompt,
    boxsep=2pt,
    left=4pt,
    right=4pt,
    top=4pt,
    bottom=4pt
]
Convert the content of a specific page from the provided PDF into LaTeX format, following these requirements:

\begin{enumerate}\itemsep2pt
    \item \textbf{Convert only the main body text}: Do not include headers, footers, page numbers, or decorative page elements.

    \item If the page is the \textbf{first page of the paper}, use \verb|\title{}| for the title, \verb|\author{}| for the author information (if any), and
\begin{verbatim}
\begin{abstract}
xxx
\end{abstract}
\end{verbatim}
for the abstract, then begin the main text using \verb|\section|.

    \item \textbf{Do not use any packages or preamble}: Output only the LaTeX code corresponding to the body content. If the beginning of the page is the continuation of a previous section, convert that text as well without omission.

    \item \textbf{Handling figures and tables}:
    \begin{itemize}\itemsep1pt
        \item Name all figure files as \verb|figure_x.pdf|, where \verb|x| is the figure's ID from the original document.
        \item If a figure contains multiple subfigures, convert it using subfigure format and name the files \verb|figure_x_1.pdf|, \verb|figure_x_2.pdf|, etc.
        \item Assign figure labels using \verb|\label{figure_x}|.
        \item Identify table IDs similarly, and use \verb|\label{table_x}| for tables.
        \item If the page contains references to figures or tables, convert them according to these rules, for example \verb|\ref{fig:figure_x}|, \verb|\ref{tab:table_x}|.
        \item You should use \verb|\begin{figure}| to describe the figure.
    \end{itemize}

    \item \textbf{Correctly convert the section structure}:
    \begin{itemize}\itemsep1pt
        \item Use \verb|\section|, \verb|\subsection|, etc., following the exact hierarchical structure of the original text.
    \end{itemize}

    \item \textbf{Handling citations}:
    \begin{itemize}\itemsep1pt
        \item Convert numeric citations like \verb|[1,2]| to \verb|\cite{1,2}|.
        \item Convert author-year citations like \verb|(Li et al., 2023; Burns et al., 2022)| into \verb|\cite{Li_2023, Burns_2022}|.
        \item If there are two authors, such as ``Pezeshkpour \& Hruschka, 2023'', convert it to \verb|\cite{Pezeshkpour_Hruschka_2023}|, ensuring the citation key reflects the authors and year.
    \end{itemize}

    \item \textbf{Incomplete content at the top or bottom of the page}: Do not modify or complete it. Keep the original meaning and convert it as is.

    \item \textbf{If the page contains part of the References section}:
    \begin{itemize}\itemsep1pt
        \item Convert all reference entries on the page into BibTeX format only (e.g., \verb|@article{Smith_2020, ...}|).
        \item Do not use \verb|\bibitem|.
        \item The BibTeX entry keys must exactly match the citation labels generated in Rule 6.
        \item If this page begins the References section, start with \verb|\section{References}|.
    \end{itemize}
\end{enumerate}

Directly output the LaTeX code, without any extra output.
\end{tcolorbox}
\label{fig:reasoning_prompt}
\end{figure}

\newpage
\section{Model Configuration}
\begin{table*}[h]
\centering
\footnotesize
\resizebox{\textwidth}{!}{%
\begin{tabular}{lllll}
\toprule
\textbf{Organization} & \textbf{Model} & \textbf{Release} & \textbf{Version} & \textbf{\# Inference Pipeline} \\
\midrule
\multicolumn{5}{c}{\emph{\textbf{Proprietary Models}}} \\
\midrule
OpenAI & GPT-5.3 & 2026-03 & \texttt{gpt-5.3-chat-latest} & API \\

\midrule
\multicolumn{5}{c}{\emph{\textbf{Open-source Multimodal Foundation Models}}} \\
\midrule
\multirow{7}{*}{Alibaba} & Qwen2.5-VL-32B & 2025-01 & \texttt{Qwen2.5-VL-32B-Instruct} & \multirow{7}{*}{vLLM} \\
& Qwen2.5-VL-7B & 2025-01 & \texttt{Qwen2.5-VL-7B-Instruct} & \\
& Qwen2.5-VL-3B & 2025-01 & \texttt{Qwen2.5-VL-3B-Instruct} & \\
& Qwen3-VL-2B & 2025-10 & \texttt{Qwen3-VL-2B-Instruct} & \\
& Qwen3-VL-4B & 2025-10 & \texttt{Qwen3-VL-4B-Instruct} & \\
& Qwen3-VL-8B & 2025-10 & \texttt{Qwen3-VL-8B-Instruct} & \\
& Qwen3-VL-32B & 2025-10 & \texttt{Qwen3-VL-32B-Instruct} & \\
\noalign{\vskip 0.5ex}\hdashline\noalign{\vskip 0.5ex}
\multirow{2}{*}{Mistral AI} & Mistral-Small-3.1 & 2025-03 & \texttt{Mistral-Small-3.1-24B} & \multirow{2}{*}{vLLM} \\
& Pixtral-12B & 2024-09 & \texttt{Pixtral-12B-2409}&\\
\noalign{\vskip 0.5ex}\hdashline\noalign{\vskip 0.5ex}
\multirow{5}{*}{Shanghai AI Lab} & InternVL3-38B & 2025-04 & \texttt{InternVL3-38B} & \multirow{5}{*}{vLLM} \\
& InternVL3-8B & 2025-04 & \texttt{InternVL3-8B} & \\
& InternVL2.5-38B & 2024-11 & \texttt{InternVL2.5-38B} & \\
& InternVL2.5-8B & 2024-11 & \texttt{InternVL2.5-8B} & \\
& InternVL2-8B & 2024-07 & \texttt{InternVL2-8B} & \\
\noalign{\vskip 0.5ex}\hdashline\noalign{\vskip 0.5ex}
\multirow{2}{*}{Microsoft} & Phi-3.5-Vision & 2024-07 & \texttt{Phi-3.5-Vision-Instruct} & \multirow{2}{*}{vLLM}\\
&Phi-4-Multimodal&2025-03&\texttt{Phi-4-Multimodal}&\\
\noalign{\vskip 0.5ex}\hdashline\noalign{\vskip 0.5ex}
\multirow{1}{*}{Llava Hugging Face} & LLaVA-OneVision-7B & 2024-09 & \texttt{llava-onevision-qwen2-7b-ov-chat-hf} & vLLM \\
\noalign{\vskip 0.5ex}\hdashline\noalign{\vskip 0.5ex}
\multirow{1}{*}{DeepSeek} & DeepSeek-OCR & 2025-10 & \texttt{DeepSeek-OCR} & vLLM \\
\noalign{\vskip 0.5ex}\hdashline\noalign{\vskip 0.5ex}
\multirow{1}{*}{Infly} & Infinity-Parser-7B & 2025-06 & \texttt{Infinity-Parser-7B} & vLLM \\
\noalign{\vskip 0.5ex}\hdashline\noalign{\vskip 0.5ex}
\multirow{1}{*}{Allenai} & olmOCR-2-7B & 2025-10 & \texttt{olmOCR-2-7B-1025} & olmOCR toolkit \\

\bottomrule
\end{tabular}
}
\caption{Details of the multimodal foundation models evaluated in our study. Models are organized by organization and aligned with performance data from the main text.}
\label{tab:model_configuration}
\end{table*}
\label{app:model_info}

\twocolumn

\clearpage
\onecolumn
\section{Error Analysis}
\label{sec:errorana}
\begin{figure*}[h]
    \centering
    \includegraphics[width = 0.98\linewidth]{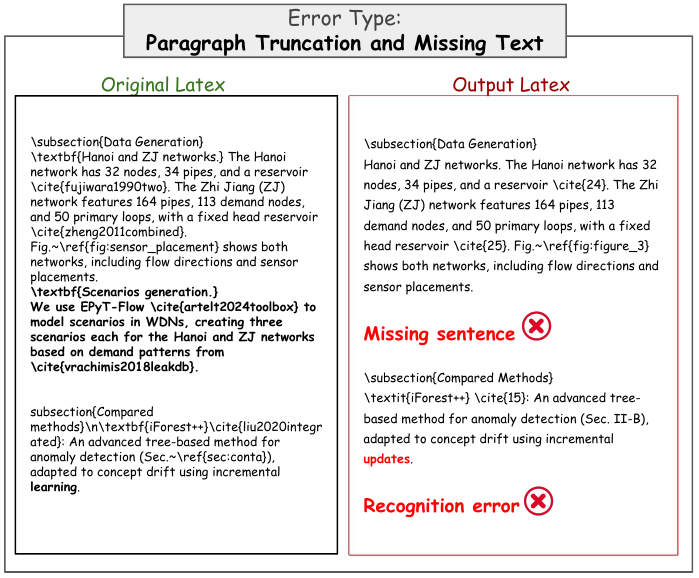}
    \caption{Example of paragraph truncation and missing text.}
    \label{fig:error1}
\end{figure*}
\twocolumn

\begin{figure*}[h]
    \centering
    \includegraphics[width = 0.98\linewidth]{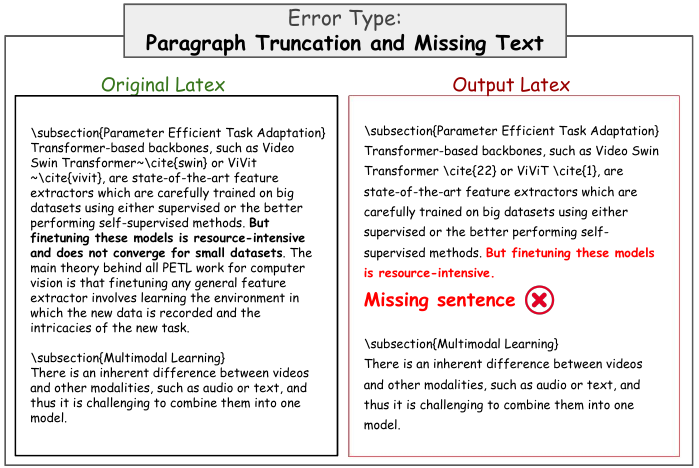}
    \caption{Example of paragraph truncation and missing text.}
\end{figure*}

\begin{figure*}[h]
    \centering
    \includegraphics[width = 0.98\linewidth]{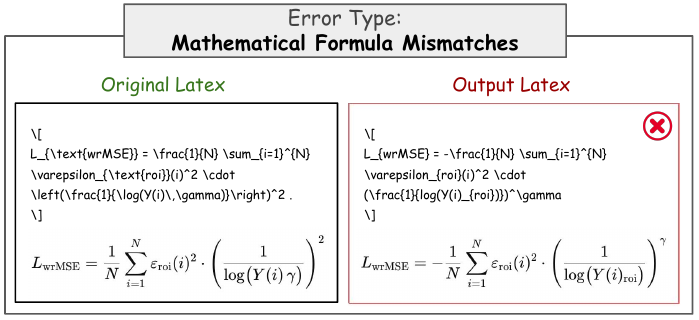}
    \caption{Example of mathematical formula mismatches.}
    \label{fig:error2}
\end{figure*}

\begin{figure*}[h]
    \centering
    \includegraphics[width = 0.98\linewidth]{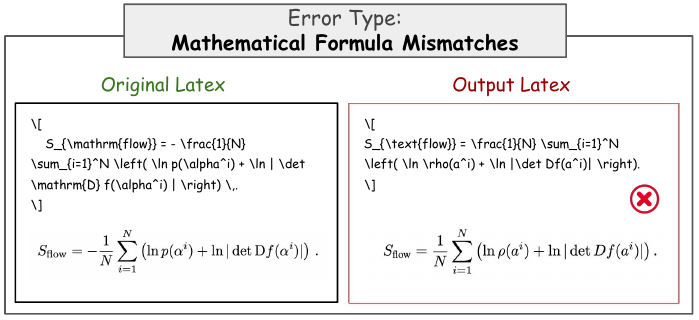}
    \caption{Example of mathematical formula mismatches.}
\end{figure*}

\begin{figure*}[h]
    \centering
    \includegraphics[width = 0.98\linewidth]{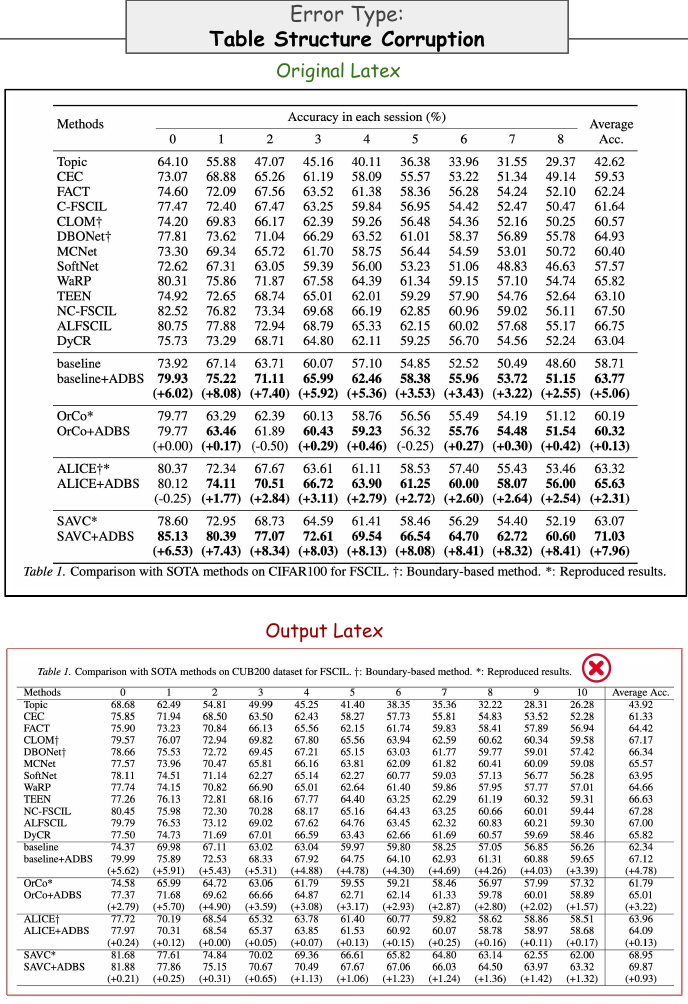}
    \caption{Example of table structure corruption.}
    \label{fig:error3}
\end{figure*}

\begin{figure*}[h]
    \centering
    \includegraphics[width = 0.98\linewidth]{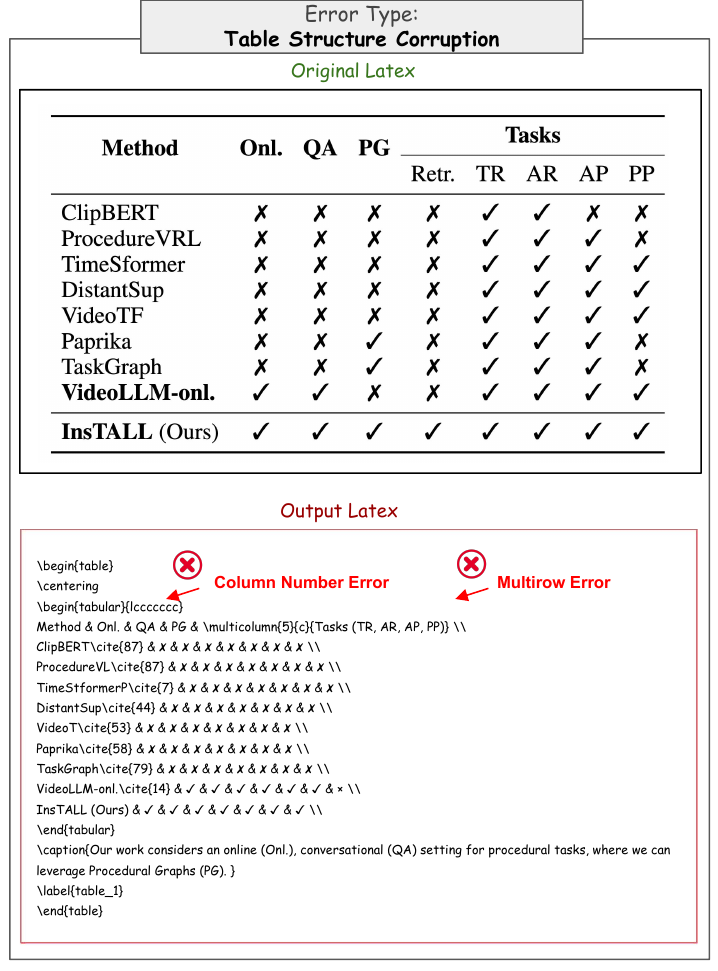}
    \caption{Example of table structure corruption.}
\end{figure*}

\begin{figure*}[h]
    \centering
    \includegraphics[width = 0.98\linewidth]{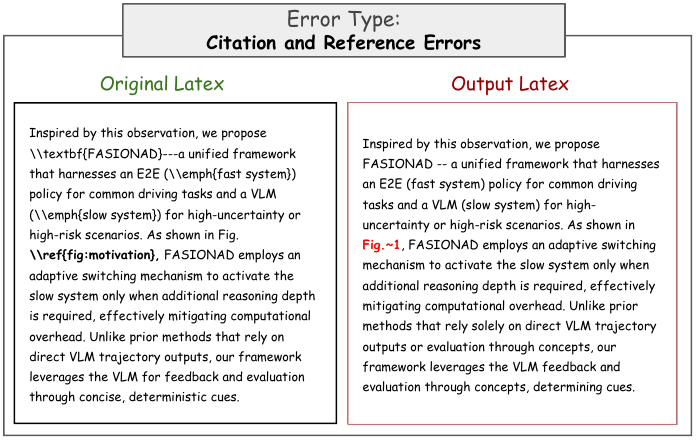}
    \caption{Example of citation and reference errors.}
    \label{fig:error4}
\end{figure*}

\begin{figure*}[h]
    \centering
    \includegraphics[width = 0.98\linewidth]{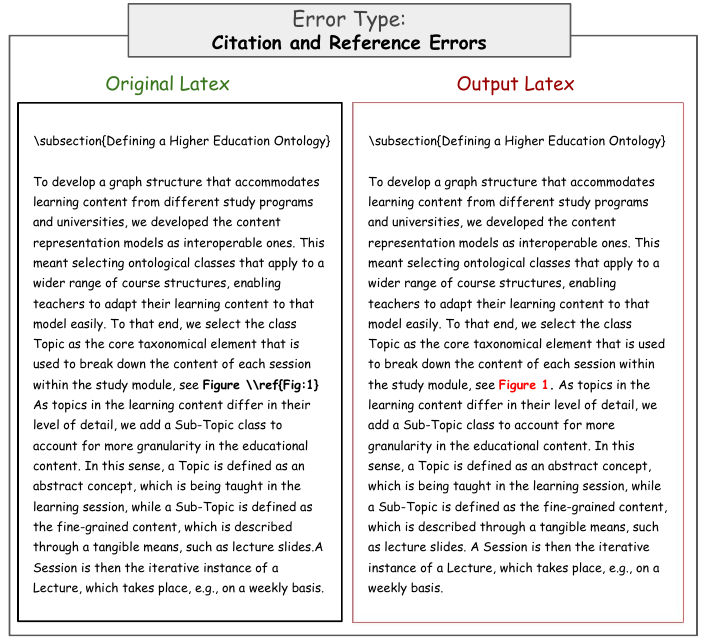}
    \caption{Example of citation and reference errors.}
\end{figure*}

\begin{figure*}[h]
    \centering
    \includegraphics[width = 0.98\linewidth]{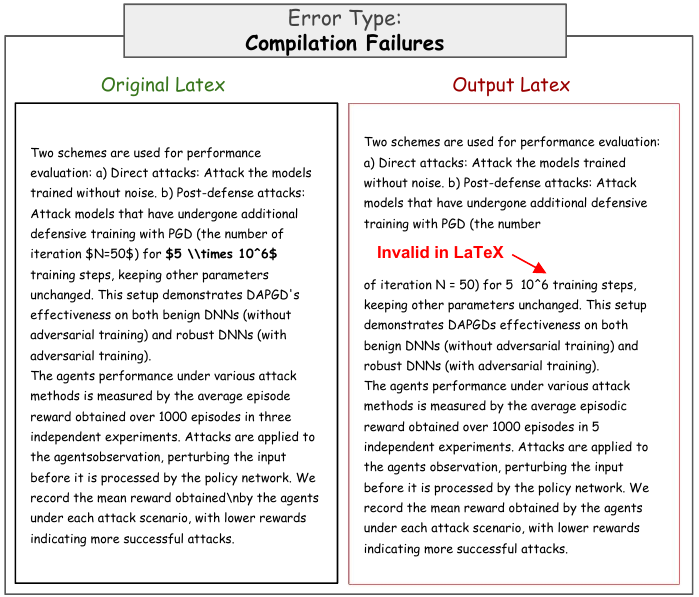}
    \caption{Example of compilation failures.}
    \label{fig:error5}
\end{figure*}

\begin{figure*}[h]
    \centering
    \includegraphics[width = 0.98\linewidth]{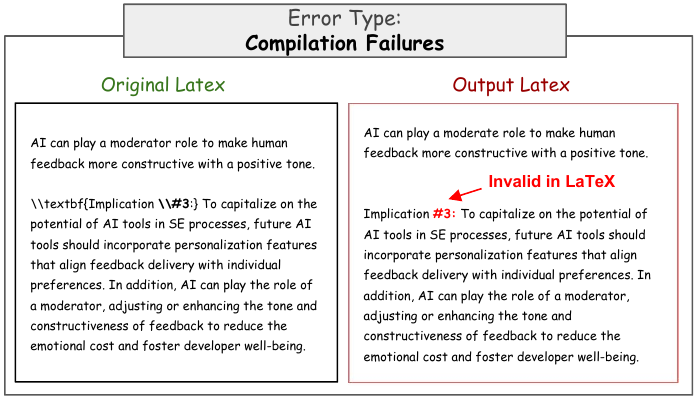}
    \caption{Example of compilation failures.}
\end{figure*}

\end{document}